\documentclass{article}
\usepackage{arxiv}
\usepackage[utf8]{inputenc} 
\usepackage[T1]{fontenc}    
\usepackage{hyperref}       
\usepackage{url}            
\usepackage{booktabs}       
\usepackage{amsfonts}       
\usepackage{nicefrac}       
\usepackage{microtype}      
\usepackage{lipsum}

\usepackage{cite}
\usepackage{amsmath,amssymb}
\usepackage{algorithmic}
\usepackage{graphicx}
\usepackage{textcomp}
\usepackage{xcolor}
\usepackage{multirow}
\usepackage{subfigure}

\title{GSIP: Green Semantic Segmentation of \\ Large-Scale Indoor Point Clouds}

\author{Min Zhang\\ Media Communications Lab\\ University of Southern California\\ Los Angeles, CA, USA\\ \texttt{zhan980@usc.edu}\\ \And Pranav Kadam\\ Media Communications Lab\\ University of Southern California\\ Los Angeles, CA, USA\\ \texttt{pranavka@usc.edu}\\ \And Shan Liu\thanks{This work was supported by Tencent Media Lab.}\\ Tencent Media Lab\\ Tencent America\\ Palo Alto, CA, USA\\ \texttt{shanl@tencent.com}\\ \And C.-C. Jay Kuo\\ Media Communications Lab\\ University of Southern California\\ Los Angeles, CA, USA\\ \texttt{cckuo@sipi.usc.edu}}

\begin{document}
\maketitle
\begin{abstract}
An efficient solution to semantic segmentation of large-scale indoor scene point clouds is proposed in this work. It is named GSIP (Green Segmentation of Indoor Point clouds) and its performance is evaluated on a representative large-scale benchmark --- the Stanford 3D Indoor Segmentation (S3DIS) dataset. GSIP has two novel components: 1) a room-style data pre-processing method that selects a proper subset of points for further processing, and 2) a new feature extractor which is extended from PointHop. For the former, sampled points of each room form an input unit. For the latter, the weaknesses of PointHop's feature extraction when extending it to large-scale point clouds are identified and fixed with a simpler processing pipeline. As compared with PointNet, which is a pioneering deep-learning-based solution, GSIP is green since it has significantly lower computational complexity and a much smaller model size. Furthermore, experiments show that GSIP outperforms PointNet in segmentation performance for the S3DIS dataset.
\end{abstract}


\section{Introduction}\label{sec:introduction}

Given a point cloud set, the goal of semantic segmentation is to label every point as one of the semantic categories. Semantic segmentation of large-scale point clouds find a wide range of real world applications such as autonomous driving in an out-door environment and robotic navigation in an in- or out-door environment. As compared with the point cloud classification problem that often targets at small-scale objects, a high-performance point cloud semantic segmentation method demands a good understanding of the complex global structure as well as the local neighborhood of each point. Meanwhile, efficiency measured by computational complexity and memory complexity is important for practical real-time systems. 

State-of-the-art point cloud classification and segmentation methods are based on deep learning. Raw point clouds captured by the LiDAR sensors are irregular and unordered. They cannot be directly processed by deep learning networks designed for 2D images. This problem was addressed by the pioneering work on the PointNet \cite{qi2017pointnet}. PointNet and its follow-ups \cite{qi2017pointnet++, jiang2018pointsift, li2018pointcnn, wang2019dynamic} achieved impressive performance in small-scale point cloud classification and segmentation tasks. A representative point cloud classification dataset is ModelNet40 \cite{wu20153d}, where each object has 1,024 or 2,048 points. 

Although the above-mentioned methods work well for small-scale point clouds, they can't be generalized to handle large-scale point cloud directly due to the memory and time constraints. Large-scale point cloud semantic segmentation methods are often evaluated on the S3DIS dataset \cite{armeni_cvpr16} for the indoor scene and the Semantic3D \cite{hackel2017semantic3d} or SemanticKITTI \cite{behley2019semantickitti} datasets for the outdoor scenes. These datasets have millions of points, covering up to $200 \times 200$ meters in 3D real-world space. In order to feed such a large amount of data to deep learning networks, it is essential to conduct pre-processing steps on the data such as block partitioning \cite{qi2017pointnet}. Recently, efforts have been made in \cite{landrieu2018large, rethage2018fully, guo2020pct, hu2020randla} to tackle with large-scale point clouds directly. For example, RandLA-Net \cite{hu2020randla} abandons block partitioning and enlarges an input unit from 4,096 points to 40,960 points. 

Green point cloud learning methods are proposed recently, e.g., \cite{zhang2020pointhop, zhang2020pointhop++, zhang2020unsupervised, kadam2020unsupervised, kadam2021r}. They are called ``green" since they have smaller model sizes and can be trained by CPU with much less time and complexity. The main saving comes from the one-pass feedforward feature learning mechanism, where no backpropagation neither supervision label is needed. The unsupervised feature learning is built upon statistical analysis of points in a local neighborhood of a point cloud set. Along this direction, we attempt to develop a green semantic segmentation solution for large-scale indoor point clouds in this work, named GSIP (Green Segmentation of Indoor Point clouds). 

GSIP has two novel components: 1) a room-style data pre-processing method and 2) an enhanced PointHop feature extractor. For the former, we compare existing data pre-process techniques and identify the most suitable data preparation method. In deep-learning methods, which are implemented on GPUs, data pre-processing is used to ensure that input units contain same number of points so that the throughput is maximized by exploiting GPU's built-in parallel processing capability. For the S3DIS dataset, a unit has 4,096 points in PointNet which adopts the block-style pre-processing while a unit has 40,960 points in RandLA-Net which adopts the view-style pre-processing. Since feature extraction for green point cloud learning runs on a CPU, we are able to relax the constraint that each input unit can have different number of points. We propose a new room-style pre-processing method for GSIP and show its advantages. For the latter, we point out some weaknesses of PointHop's feature extraction when extending it to large-scale point clouds and fix them with a simpler processing pipeline.

The proposed GSIP method can be stated below. A raw point cloud set is voxel-downsampled with a fixed grid size. All down-sampled points in one room form one unit whose point number ranges from 10K to 200K. Then, feature of points in each unit are fed into an unsupervised feature extractor to obtain point-wise features at various hops. Finally, features are fed to the XGBoost classifier \cite{chen2015xgboost} for point-wise classification, i.e., semantic segmentation. We conduct experiments on the S3DIS dataset \cite{wu20153d} and show that, besides a smaller model size and lower computational complexity, GSIP outperforms PointNet in segmentation performance.  Our work has the following three main contributions:
\begin{enumerate}
\item We compare existing data pre-processing methods for point cloud semantic segmentation and identify the most suitable data preparation method. 
\item We propose a lightweight and efficient unsupervised feature extraction method in GSIP. 
\item We demonstrate the performance, efficiency and model size gains of GSIP over PointNet on the S3DIS dataset. 
\end{enumerate}

\section{Related Work}\label{sec:review}

\subsection{Traditional Methods}

Traditional features for 3D point clouds are extracted using a hand-crafted solution. It does not involve any training data but relies on local geometric properties of points. FPFH \cite{rusu2009fast} and SHOT \cite{tombari2010unique} are exemplary feature descriptors. Semantic segmentation \cite{hackel2016fast, landrieu2017structured} is formulated as a point-wise classification problem. Each point is described by one or multiple feature descriptors, and its extracted features are concatenated to form a feature vector. Then, the feature vector is fed into a classifier such as the Support Vector Machine (SVM) \cite{mallet2011relevance} and the Random Forest (RF) \cite{chehata2009airborne}. The classification stage does need representative training data for classifier's training. In the inference stage, the classifier will assign a class label to each point of the target point cloud set. 

\subsection{Deep Learning Methods}\label{subsec:deep}

One pioneering work in deep learning methods is PointNet
\cite{qi2017pointnet}.  It adopts multi-layer perceptrons (MLPs) and max pooling to learn point-wise features. Yet, PointNet fails to capture the local structure of a point. Its follow-ups \cite{qi2017pointnet++, jiang2018pointsift, li2018pointcnn, wang2019dynamic} attempt to capture the local information of a point to learn a richer context. For example, PointNet++ \cite{qi2017pointnet++} applies PointNet to the local region of each point, and then aggregates local features in a hierarchical architecture. DGCNN \cite{wang2019dynamic} exploits another idea to learn better local features. It uses point features to build a dynamical graph and updates neighbor regions at every layer. A dynamic graph can capture better semantic meaning than a fixed graph. PointCNN \cite{li2018pointcnn} learns an $\chi$-transformation from an input point cloud to get weights of neighbors of each point and permute points based on a latent and potentially canonical order. To make local features invariant to permutations, PointSIFT \cite{jiang2018pointsift} discards the max pooling idea but designs an orientation-encoding unit to encode the eight orientations, which is inspired by the well-known 2D SIFT descriptor \cite{lowe2004distinctive}. RandLA-Net \cite{hu2020randla} learns important local features through the attention mechanism. A shared MLP followed by the softmax operation is used to compute attention scores. Then, attention-weighted local features are summed together. 

\subsection{Green Learning Methods} \label{sec:explainable}

\begin{table}[!t]
\newcommand{\tabincell}[2]{\begin{tabular}{@{}#1@{}}#2\end{tabular}}
\centering
\caption{Comparison of traditional, deep learning (DL) and green learning (GL) methods.}
\label{tab:related_compare}
\renewcommand\arraystretch{1.3}
\begin{tabular}{|c|c|c|c|} \hline
   Feature & Traditional & DL & GL \\ \hline\hline
   Data eagerness & low & high & middle \\ \hline
   Supervision & low & high & middle \\ \hline 
   Model size  & small & large & middle \\ \hline 
   Time complexity  & low & high & middle \\ \hline 
   Interpretation  & easy & hard & easy \\ \hline 
   Performance  & poor & good & good \\ \hline 
\end{tabular}
\end{table}

Green point cloud learning has been introduced in a sequence of recent publications \cite{zhang2020pointhop, zhang2020pointhop++, zhang2020unsupervised, kadam2020unsupervised, kadam2021r}. Its theoretical foundation is successive subspace learning (SSL) \cite{kuo2016understanding, kuo2018data, kuo2019interpretable, chen2018saak, chen2020pixelhop, chen2020pixelhop++, yang2021pixelhop}, which applies to 2D images as well as 3D point clouds. SSL helps reduce the model size and computation complexity, and offers mathematical transparency. As compared with traditional and deep learning methods. We compare traditional, deep learning and green
learning methods qualitatively in several aspects in Table
\ref{tab:related_compare}. As shown in the table, green point cloud learning methods have several advantages:
\begin{enumerate}
\item data-driven but not data-eager (i.e. robust with less data),
\item no supervision needed for feature extraction,
\item good generalization ability, where extracted feature can be used
for multi-tasking such as object classification, part segmentation and
registration,
\item mathematically interpretable,
\item smaller model sizes and lower computation complexity.
\end{enumerate}
Apart from point cloud processing, the green learning technology has been successfully applied to other tasks such as face gender classification \cite{rouhsedaghat2020facehop}, deepfake image detection \cite{chen2021defakehop}, and image anomaly detection \cite{zhang2021anomalyhop}. 

\subsection{Large-scale Point Cloud Learning}\label{sec:large}

Point-based deep learning methods as reviewed in Sec. \ref{subsec:deep} are limited to small-scale point clouds. For large-scale point cloud learning, SPG \cite{landrieu2018large} builds super graphs composed by super points in a pre-processing step. Then, it learns semantics for each super point rather than a point. FCPN \cite{rethage2018fully} and PCT \cite{guo2020pct} combine voxelization and point-based networks to process large-scale point clouds. However, graph construction and voxelization are computationally heavy so that these solutions are not suitable for mobile or edge devices. Recently, RandLA-Net \cite{hu2020randla} revisits point-based deep learning methods. It replaces the farthest point sampling (FPS) method with random sampling (RS) to save time and memory cost. In this way, the number of points that can be processed one time is increased by 10 times. However, RS may discard key features, so it is not as accurate as FPS. To address this problem, a new local feature aggregation module is adopted to capture complex local structures. 

\section{Proposed GSIP Method}\label{sec:method}

\begin{figure*}[!t]
\centering
\includegraphics[scale=0.47]{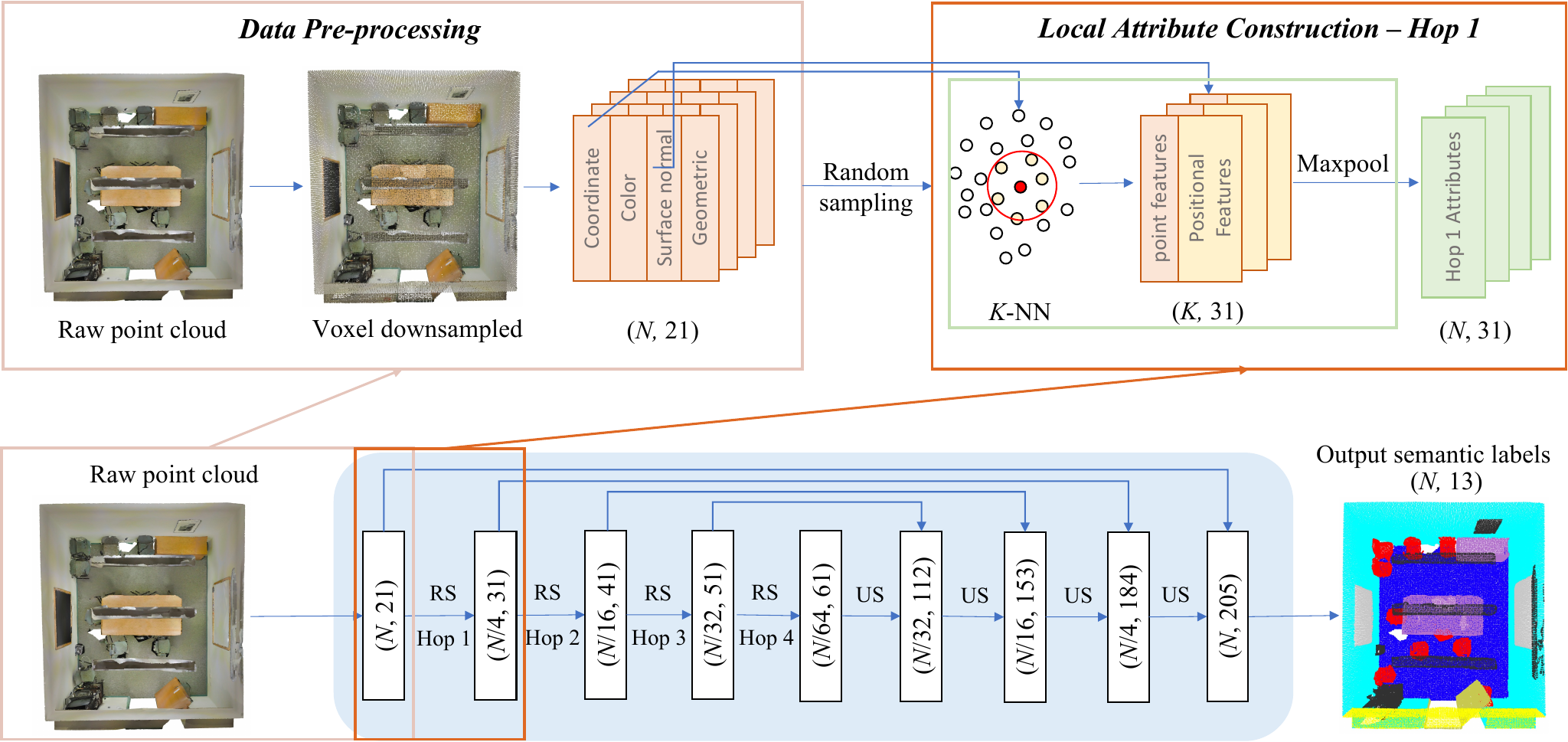}
\caption{An overview of the proposed GSIP method, where the upper left block shows the data pre-processing, the upper right block shows the local attribute construction process and the lower block shows the encoder-decoder architecture for large-scale point cloud semantic segmentation.}
\label{fig:1} 
\end{figure*}

An overview of the proposed GSIP method is given in Fig. \ref{fig:1}. A raw point cloud set is first voxel-downsampled, where each downsampled point has 9D attributes (i.e., XYZ, RGB, and normalized XYZ). Each sampled point is augmented with 12D additional attributes. They include: surface normals (3D), geometric features (6D, planarity, linearity, surface variation, etc.) \cite{hackel2016fast} and HSV color space (3D, converted from RGB). Surface normals and geometric features are commonly used in traditional point cloud processing.  As compared with the XYZ coordinates, they describe local geometric patterns and can be easily computed. Here, we include them as additional features to the XYZ coordinates. For points that have the same geometric pattern but belong to different objects (e.g., wall and blackboard), the RGB plus HSV color spaces work better than RGB alone. Consequently, each point has 21D input features. The point-wise features are not powerful enough since they are localized in the space. To obtain more powerful features, we need to group points based on different neighborhood sizes (e.g., points in a small neighborhood, a mid-size neighborhood and a large neighborhood). By borrowing the terminology from graph theory, we call these neighborhoods hops, where hop 1 denotes the smallest neighborhood size. To carry out the semantic segmentation task, we need to extract features at various hops, which is achieved by an unsupervised feature extractor. The feature extractor adopts the encoder-decoder architecture. It has four encoding hops followed by four decoding hops. Finally, a classifier is trained and used to classify each point into a semantic category based on its associated hop features.

\subsection{Data Pre-processing}

Data pre-processing is used to prepare the input data in proper form so that they can be fed into the learning pipeline for effective processing. Although the generic principle holds for any large-scale point clouds, the implementation detail depends on the application context. Data pre-processing techniques targeting at large-scale indoor scene point clouds are discussed here. Specifically, we use the S3DIS dataset \cite{armeni_cvpr16}, which is a subset of the Stanford 2D-3D-Semantics dataset \cite{armeni2017joint}, as an example. S3DIS is one of the benchmark datasets for point cloud semantic segmentation. It contains point clouds scanned from 6 indoor areas with 271 rooms. There are 13 object categories in total, including ceiling, floor, wall, door, etc. Each point has 9 dimensions, i.e., XYZ, RGB and normalized XYZ. Data pre-processing techniques can be categorized into the following three styles. 

\begin{figure}[t!]
\centering
\begin{minipage}[b]{0.98\textwidth}
  \centering
  \centerline{\includegraphics[scale=.45]{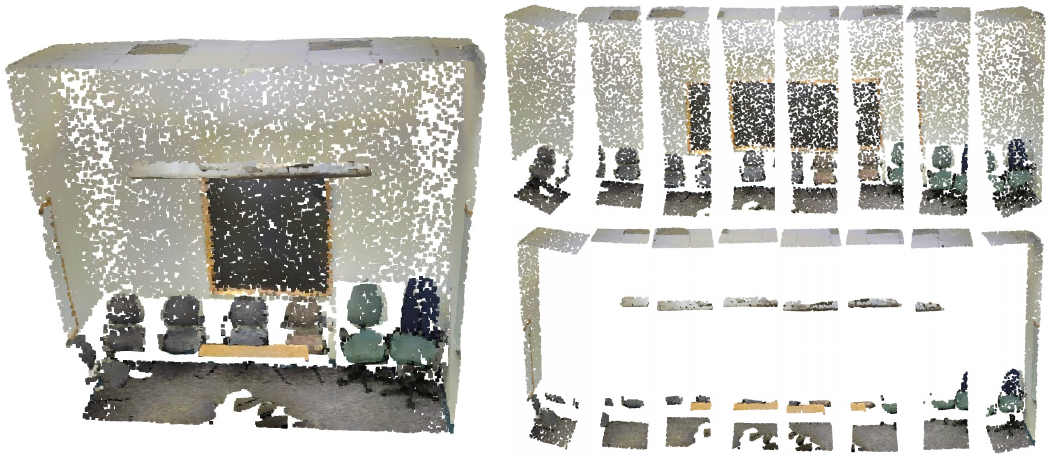}}
  \centerline{(a) Block style}\medskip
\end{minipage}
\begin{minipage}[b]{0.98\textwidth}
  \centering
  \centerline{\includegraphics[scale=.45]{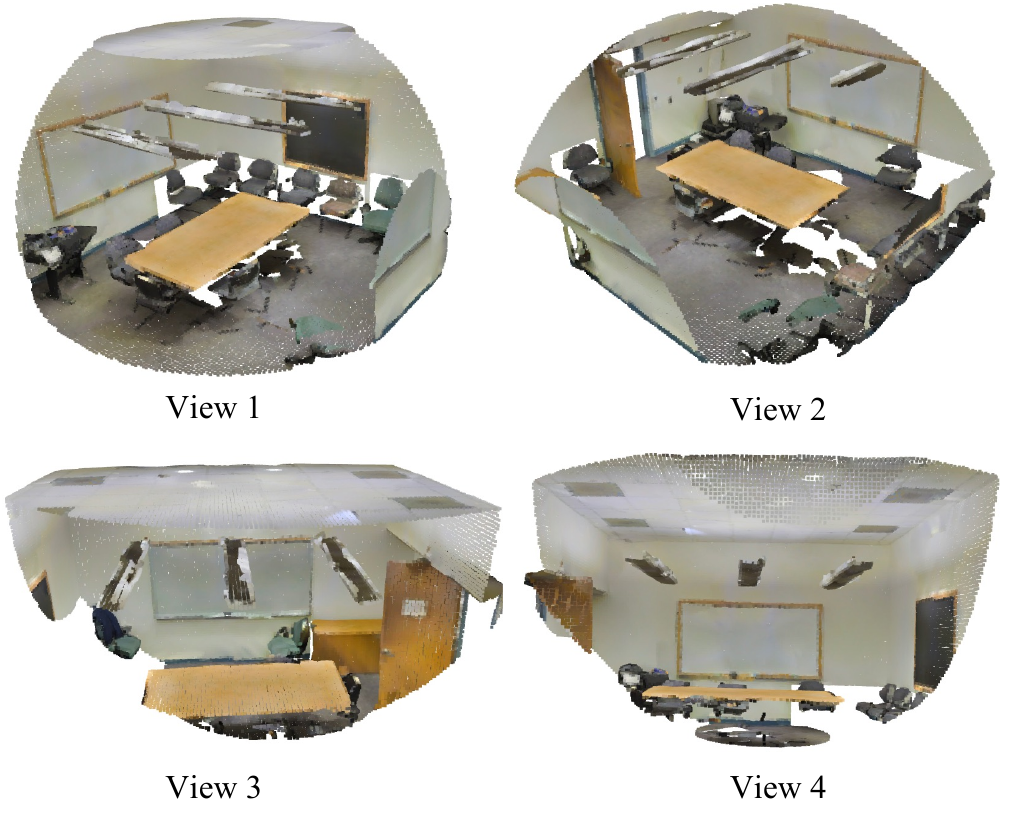}}
  \centerline{(b) View style}\medskip
\end{minipage}
\begin{minipage}[b]{0.98\textwidth}
  \centering
  \centerline{\includegraphics[scale=.45]{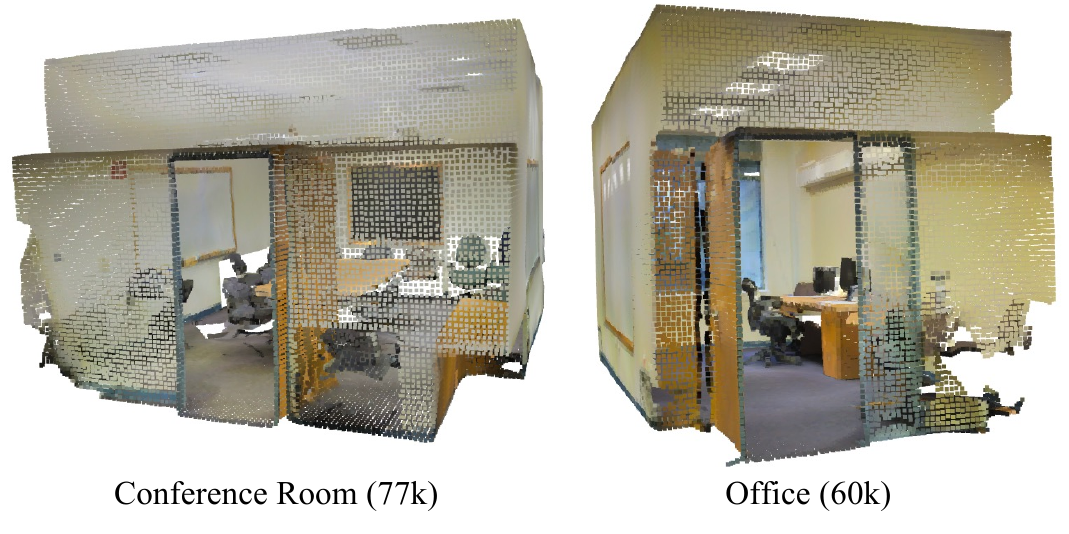}}
  \centerline{(c) Room style}\medskip
\end{minipage}
\caption{Comparison of three data pre-processing methods.}\label{fig:2}
\end{figure}

\textbf{Block Style.} Block partitioning was proposed by PointNet and adopted by its follow-ups. The 2D horizontal plane of a room is split into $1 \times 1$ meter blocks while its 1D vertical direction is kept as it is to form an input unit as shown in Fig. \ref{fig:2}(a). Each unit contains 4,096 points, which is randomly sampled from its raw block. In the inference stage, 4,096 points of each unit are classified and, then, their predicted labels are propagated to their neighbors. Typically, the k-fold strategy is adopted for train and test. For example, if area 6 is selected as the test area, the remaining 5 areas are used for training. Under this setting, the block-style data pre-processing has 20,291 training units and 3,294 testing units, where each unit has 4,096 points. 

\textbf{View Style.} View-style data pre-processing was adopted by RandLA-Net, but not explained in the paper. We get its details from the codes and describe its process below. It first partitions the 3D space of a room into voxel grids and randomly selects one point per voxel. For instance, the first conference room of area 1 has about one million points. One can obtain 70k points by setting the voxel grid size to 0.04, around 7.7\% points are kept in this example. This is a commonly used point downsampling procedure. Then, it iteratively selects a fixed number of points generating input units to facilitate GPU parallel processing. For initialization, it randomly selects a room and a point in the room as the reference point. Then, it finds the 40,960 nearest points around the reference point by $K$-NN algorithm to form the first unit. This process is repeated to get a sequence of input units until it reaches the target unit number. For the S3DIS dataset, the target training and test unit numbers for each fold are set to 3K and 2K, respectively. To reduce the overlapping of different units, it assigns a possibility to every point randomly at the beginning and updates the possibilities of the selected points in each round as inversely proportional to their distances to the reference point. Thus, unselected points will be more likely to be chosen as the next reference point. Four examples are shown in Fig. \ref{fig:2}(b) to visualize points inside the same unit. We see that they offer certain views to a room. 
  
There are however several problems for the above two data pre-processing methods. First, they do not have a global view of the whole room, resulting in inaccurate nearest neighbor search in unit boundaries. Second, the view-style method generates 3K units for 200 rooms (i.e., 15 units per room on the average) in the training. There are redundant points in the intersection of views of the same room. The total number of training points increases from 80M $(\approx 20,291 \times 4,096)$ of the block style to 120M $(\approx 3,000 \times 40,960)$ of the view style. We may ask whether it is essential to have so many training points. 

\textbf{Room Style.} It is desired to increase input scale and reduce the total number of training points while keeping the same segmentation performance. To achieve this goal, we propose a room-style pre-processing method and adopt a flexible feature extraction pipeline, which will be discussed in Sec. \ref{subsec:feature}. The most distinctive aspect of the room style is that each unit can have a different number of points. Thus, we include all points in a room in one unit and call it room-style pre-processing. This is a possible solution since our point cloud feature extractor can be implemented in the CPU effectively. By relaxing the constraint that all units have the same number of points demanded by the GPU implementation, the data pre-processing problem can be greatly simplified. By following the first step of RandLA-Net's pre-processing, we downsample raw points by voxel downsampling method with a fixed grid size. Afterwards, the number of points for each room ranges from 10K to 200K. By taking rooms in areas 1-5 as training and rooms in area 6 as testing as an example, we have 224 rooms (or 224 units) for training and 48 rooms (or 48 units) for testing. The training and testing sets contain 15M and 2M points, respectively. The total number of training points is much smaller than those of the block-style and the view-style methods while the input scale is much larger.

\subsection{Feature Extractor and Classifier}\label{subsec:feature}

Points in a room-style input unit are fed into a feature extractor to obtain point-wise features as shown in Fig. \ref{fig:1}. The upper right block shows the local attribute construction process and the lower block shows the encoder-decoder architecture for large-scale point cloud semantic segmentation. It is developed upon our previous work PointHop. The reason for developing new feature learner is due to several shortcomings when extending PointHop from small-scale to large-scale point cloud learning. It is detailed below. 

\textbf{PointHop and PointHop++ Feature Extractor.} PointHop \cite{zhang2020pointhop} and PointHop++ \cite{zhang2020pointhop++} are unsupervised feature extractors proposed for small-scale point cloud classification. They have been successfully applied to joint point cloud classification and part segmentation \cite{zhang2020unsupervised} and point cloud registration \cite{kadam2020unsupervised, kadam2021r}. PointHop constructs attributes from local to global by stacking several hops, covering small-, mid- and long-range neighborhoods. In each hop, the local neighborhood of each point are divided by eight octants using the the 3D coordinates of local points. Then, point attributes in each octant are aggregated and concatenated to form a local descriptor, which keeps more information than naive max pooling. Due to the fast dimension growing, the Saab transform \cite{kuo2019interpretable}, which is a variant of Principal Component Analysis (PCA) \cite{wold1987principal}, is used for dimension reduction is used for dimension reduction. Between two consecutive units, FPS is used to downsample the point cloud to increase the speed as well receptive field of each point. PointHop++ \cite{zhang2020pointhop++} is an extension of PointHop. It has a lower model size and training complexity by leveraging the observation that spectral features are uncorrelated after PCA. Thus, we can conduct PCA to each spectral feature independently, which is called the channel-wise Saab transform. 

There are three shortcomings for the pipeline used in PointHop and PointHop++ when it is ported to large-scale point cloud data. First, the computational efficiency is limited by FPS between two consecutive hops. Second, the memory cost of eight-octant partitioning and grouping is high. Third, the covariance matrix computation in the Saab transform is computationally intensive with a higher memory cost. To address them, we propose several changes. 

\textbf{Proposed Feature Extractor.} As shown in Fig. \ref{fig:1} (enclosed by the orange block), the new processing module contains random sampling (RS), $K$-NN, relative point positional encoding \cite{hu2020randla}, max pooling, and point feature standardization. First, we use RS rather than FPS between hops for faster computation of large-scale point clouds. Second, to reduce the memory cost of the eight-octant partitioning and grouping, we adopt max pooling. Since the feature dimension remains the same with max pooling prevents, no dimension reduction via the Saab transform is needed, which further helps save memory and time cost. However, max pooling may drop important information occasionally. To address it, we add relative point positional encoding before max pooling to ensure that point features are aware of their relative spatial positions. Specifically, for point $p_i$ and its $K$ neighbors $\{p_i^1, \cdots, p_i^k, \cdots, p_i^K\}$, the relative point position $r_i$ of each neighbor $p_i^k$ is encoded as 
\begin{equation}
r_i = p_i \oplus p_i^k \oplus (p_i - p_i^k) \oplus \|p_i - p_i^k\|,
\end{equation}
where $\oplus$ denotes concatenation, and $\| \cdot \|$ is the Euclidean distance. We will show that the new pipeline is much more economic than PointHop, PointHop++ and deep-learning methods in terms of memory consumption and computational complexity in Sec. \ref{sec:experiments}.

\textbf{Classifier.} The S3DIS dataset is highly imbalanced in point labels. Among the 13 object categories, the ceiling, floor and wall three classes take around 75\% of the data. We adopt the XGBoost classifier \cite{chen2015xgboost} to help reduce the influence of imbalanced data. Other classifiers such as Linear Least Square Regression, Random Forest and Support Vector Machine (SVM) demand higher computational speed and memory cost in the large-scale point cloud segmentation problem. Overall, XGBoost can handle the large-scale point cloud data with good performance, fast speed and lower memory requirement.

\section{Experiments}\label{sec:experiments}

{\bf Experimental Setup.} We adopt the following setting to evaluate the semantic segmentation performance for the S3DIS dataset. The grid size is 0.04 in voxel-based downsampling. The feature extractor has 4 hops. We set $K=64$ in $K$-NN search. Some methods may choose smaller $K$ to save computational complexity, i.e., 16, 20 and 32, but our method can afford a larger $K$. Because the input is a large-scale indoor scene, a larger $K$ will lead to a larger receptive field which helps learn the structure of the scene. Thus, we choose 64 here. The subsampling ratios between two consecutive hops are 0.25, 0.25, 0.5, 0.5, respectively. Three nearest neighbors' features are used to interpolate in the upsampling module. For example, to upsample from $N/64$ points to $N/32$ points (see Fig.1), we first find the three nearest points in the $N/64$ point set for each point in the $N/32$ point set. Then, we average the features of the three points. In this way, we get the point features for $N/32$ points. The output features are truncated to 32-bit before fed into the XGBoost classifier. The standard 6-fold cross validation is used in the experiment, where one of the six areas is selected as test area in each fold. By following \cite{qi2017pointnet}, we consider two the evaluation metrics: the mean Intersection-over-Union (mIoU) and the Overall Accuracy (OA) of the total 13 classes. 

{\bf Comparison of Data Pre-processing Methods.} The statistics of the S3DIS dataset using three data pre-processed methods are compared in Table \ref{tab:1}. The proposed room-style method has more points in each input unit, i.e., input scale, ranging from 10K to 200K. We also compare the data size when the training data are collected from areas 1-5 and the test data are collected from area 6. The total number of training points of the room-style method is 18.75\% of the block-style method and 12.5\% of the view-style method. Points inside each unit of the three methods are visualized in Fig. \ref{fig:2}, which includes a conference room and an office from area 1. The block-style method loses the structure of a complete room. As to the view-style method, the four views of a conference room overlap a lot with each other, producing significant redundancy. The units obtained by the room-style method look more natural. They offer a view of the whole room while keeping a small data size. 

\begin{table}[h]
\newcommand{\tabincell}[2]{\begin{tabular}{@{}#1@{}}#2\end{tabular}}
\centering
\caption{Comparison of data statistics of three pre-processing methods.}
\label{tab:1}
\renewcommand\arraystretch{1.3}
\begin{tabular}{|c|c|c|c|c|} \hline
   \multicolumn{2}{|c|}{Method} & Block & View & Room \\ \hline\hline
   \multicolumn{2}{|c|}{Input Scale ($\times 10^3$)} & 4 & 40 & 10-200 \\ \hline \hline
   \multirow{2}*{\tabincell{c}{Total Data Size \\ ($\times 10^6$)}} & Train  & 80 & 120 & 15 \\ \cline{2-5}
   & Test  & 10 & 80 & 2 \\ \hline 
\end{tabular}
\end{table}

\begin{figure}[!ht]
\centering
\includegraphics[scale=0.45]{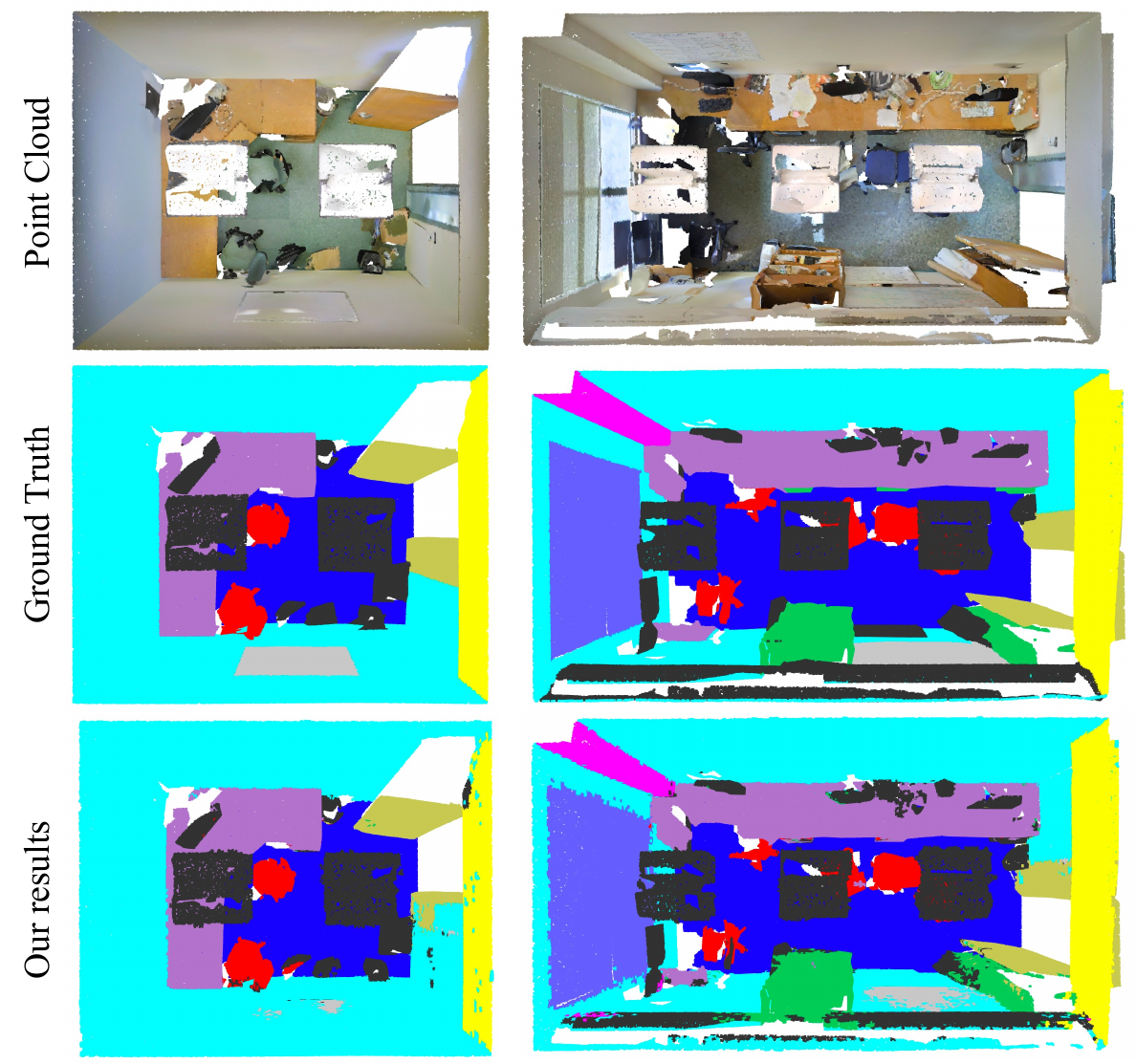}
\caption{Qualitative results of the proposed GSIP method.}
\label{fig:3} 
\end{figure}

\textbf{Semantic Segmentation Performance.} We compare the semantic segmentation performance on the S3DIS dataset of PointNet and the proposed GSIP in Table \ref{tab:2}, where the better results are shown in bold. It is common to use area 5 as the test. Thus, we show performance for area 5 as well as the 6-fold. As shown in the table, GSIP outperforms PointNet in mIoU by 2.7\% and 0.9\% for area 5 and 6-fold, respectively. The cross validation results of all 6 areas of GSIP are shown in Table \ref{tab:3}. We see that area 6 is the easiest one (64.5\% mIoU and 86.5\% OA) while area 2 is the hardest one (31\% and 68.8\%). The mIoU and OA over 13 classes averaged by the 6-fold are 48.5\% and 79.8\%, respectively. Visualization of GSIP's segmentation results and the ground truth of two room in area 6 is given in Fig. \ref{fig:3}. The ceiling is removed to show inner objects clearly.

It is worthwhile to point out that some categories have extremely low mIoU (close to 0\%) in more than 2 areas. For example, sofa got 1.7\%, 6.4\%, 4.4\%, and 3.4\% mIoU in areas 2, 3, 4 and 5. This is attributed to data imbalance. The data imbalance problem is commonly seen in large-scale segmentation tasks. For example, in a regular room, it is highly possible that more points are from the wall, ceiling and floor, while less points from chairs, desks, and other small objects. The beam, column, sofa, and board are even less common. Without seeing enough samples during training, the decrease in prediction performance is valid.

\begin{table*}[!ht]
\centering
\caption{Comparison of semantic segmentation performance (\%) for S3DIS.}
\label{tab:2}
\renewcommand\arraystretch{1.3}
\setlength{\tabcolsep}{0.6mm}{
\begin{tabular}{|c|c|c|c|c|c|c|c|c|c|c|c|c|c|c|c|c|} \hline
    & Method & OA & mIoU & ceiling & floor & wall & beam & column & window & door & table & chair & sofa & bookcase & board & clutter \\ \hline \hline
   \multirow{2}*{Area 5} & PointNet & - & 41.1 & 88.0 & \bf{97.3} & 69.8 & 0.05 & 3.90 & \bf{42.3} & 10.8 & 58.9 & 52.6 & \bf{5.90} & 40.3 & \bf{26.4} & 33.2 \\ \cline{2-17}
   & GSIP & \bf{79.9} & \bf{43.8} & \bf{89.2} & 97.0 & \bf{72.2} & \bf{0.10} & \bf{18.4} & 37.3 & \bf{22.5} & \bf{64.3} & \bf{59.5} & 3.40 & \bf{47.2} & 22.9 & \bf{35.7} \\ \hline \hline
   \multirow{2}*{6-fold} & PointNet & 78.5 & 47.6 & 88.0 & 88.7 & 69.3 & \bf{42.4} & 23.1 & \bf{47.5} & 51.6 & 54.1 & 42.0 & 9.60 & 38.2 & \bf{29.4} & 35.2 \\ \cline{2-17}
   & GSIP & \bf{79.8} & \bf{48.5} & \bf{91.8} & \bf{89.8} & \bf{73.0} & 26.3 & \bf{24.0} & 44.6 & \bf{55.8} & \bf{55.5} & \bf{51.1} & \bf{10.2} & \bf{43.8} & 21.8 & \bf{43.2} \\ \hline
\end{tabular}}
\end{table*}

\begin{table}[!ht]
\centering
\caption{Class-wise semantic segmentation performance (\%) of GSIP for S3DIS.}
\label{tab:3}
\renewcommand\arraystretch{1.3}
\begin{tabular}{|c|c|c|c|c|c|c||c|} \hline
   & 1 & 2 & 3 & 4 & 5 & 6 & mean \\ \hline \hline
   ceiling  & 91.9 & 88.9 & 95.2 & 89.9 & 89.2 & 95.6 & 91.8 \\ 
   floor & 93.7 & 58.5 & 97.7 & 95.2 & 97.0 & 96.8 & 89.8 \\ 
   wall & 71.5 & 70.9 & 73.3 & 72.5 & 72.2 & 77.5 & 73.0 \\ 
   beam & 21.7 & 5.90 & 64.9 & 0.20 & 0.10 & 65.2 & 26.3 \\ 
   column & 38.7 & 12.8 & 20.9 & 15.9 & 18.4 & 37.0 & 24.0 \\ 
   window & 77.5 & 18.3 & 39.7 & 15.1 & 37.3 & 79.5 & 44.6 \\ 
   door & 71.5 & 46.0 & 69.0 & 51.7 & 22.5 & 73.8 & 55.8 \\ 
   table & 61.7 & 23.3 & 62.5 & 49.8 & 64.3 & 71.3 & 55.5 \\ 
   chair & 49.4 & 21.8 & 60.4 & 49.3 & 59.5 & 66.3 & 51.1 \\ 
   sofa & 20.4 & 1.70 & 6.40 & 4.40 & 3.40 & 24.7 & 10.2 \\ 
   bookcase & 41.2 & 22.8 & 58.1 & 34.4 & 47.2 & 58.9 & 43.8 \\ 
   board & 29.4 & 5.00 & 22.8 & 14.2 & 22.9 & 36.7 & 21.8 \\ 
   clutter & 46.6 & 27.6 & 51.3 & 42.4 & 35.7 & 55.7 & 43.2 \\ \hline\hline
   \bf{mIoU} & 55.0 & 31.0 & 55.6 & 41.2 & 43.8 & 64.5 & 48.5 \\ 
   \bf{OA} & 81.1 & 68.8 & 83.9 & 78.8 & 79.9 & 86.5 & 79.8 \\ \hline
\end{tabular}
\end{table}

\textbf{Comparison of Model and Time Complexities.} We compare the model size and training time complexity of GSIP and PointNet in Table \ref{tab:4}. PointNet takes 22 hours to train in a single GeForce GTX TITAN X GPU. GSIP takes around 40 minutes for feature extraction with a Intel(R)Xeon(R) CPU and 20 minutes to train the XGBoost classifier with 4 GeForce GTX TITAN X GPUs. The total training time is around 1 hour for each fold, which is much faster than PointNet. If we use a single GPU for XGBoost training, the overall training time is still significantly less than PointNet. To justify our claim, we calculate the time complexity of algorithm theoretically in terms of big $O$. The proposed method is composed of data pre-processing, feature extractor and classifier. For each sample with $N$ points,
\begin{enumerate}
\item data pre-processing: voxel downsampling takes $O(NlogN)$; geometric feature calculation takes $O(NK^2)$, $K$ is the number of nearest points;
\item feature extraction: random sampling takes $O(1)$; $K$-NN search takes $O(KND)$, $D$ is feature dimension; relative point position encoding takes $O(KND)$; max pooling takes $O(N)$. To sum up, the feature extractor costs $O(KND)$;
\item XGBoost classifier: the XGBoost classifier’s complexity can be found in their paper, which is not our focus. 
\end{enumerate}
For the algorithms involved, brute force time complexities are used for calculation. There are better implementations that can optimize the complexity, which are not considered here.

As to the model size, GSIP has 24K parameters while PointNet has 900K parameters. PointNet is an end-to-end supervised method, which uses some fully connected layers at the end of their pipeline to predict point labels. We do not break the pipeline into the feature extractor and the classifier for comparison. The parameters mainly come from the XGBoost classifier, which has 128 trees and the maximum depth of a tree is 6. For each tree, there are 2 parameters per intermediate node and 1 parameter per leaf node. The feature extractor only has 2 parameter in each hop, mean and standard deviation, in point feature standardization.

\begin{table}[!ht]
\newcommand{\tabincell}[2]{\begin{tabular}{@{}#1@{}}#2\end{tabular}}
\centering
\caption{Comparison of time and model complexities.}
\label{tab:4}
\renewcommand\arraystretch{1.3}
\begin{tabular}{|c|c|c|c|c|} \hline
   \multicolumn{2}{|c|}{Method} & Device & Training time & Parameter No. \\ \hline \hline
   \multicolumn{2}{|c|}{PointNet} & GPU & 22 hours &  900169 \\ \hline\hline
   \multirow{2}*{GSIP} & Feature & CPU & 40 minutes & 8 \\ \cline{2-5}
   & XGBoost & GPU & 20 minutes & 24320 \\ \hline
\end{tabular}
\end{table}

\textbf{Other Comparisons.} As discussed in Sec. \ref{subsec:feature}, the new feature extractor is more effective than the PointHop feature extractor in terms of computational and memory efficiency. It is worthwhile to compare the segmentation performance of the two to see whether there is any performance degradation. We compare the effectiveness of the new feature extractor and the PointHop feature extractor under the same GSIP framework for the S3DIS dataset in Table \ref{tab:5}. Their performance is comparable as shown in the table. Actually, the new one achieves slightly better performance. Furthermore, we compare the quantization effect of extracted features before feeding them into the XGBoost classifier in Table \ref{tab:6}. We see little performance degradation for features to be quantized to 16 or 32 bits. Thus, we can save computation and memory for classifier training and testing by using 16-bit features. 

\begin{table}[!ht]
\centering
\caption{Performance comparison of two feature extractors (\%).}\label{tab:5}
\renewcommand\arraystretch{1.3}
\begin{tabular}{|c|c|c|} \hline
   Method & mIoU & OA \\ \hline \hline
   GSIP & 48.5 & 79.8 \\ \hline
   PointHop & 47.9 & 79.1 \\ \hline
\end{tabular}
\end{table}

\begin{table}[!ht]
\centering
\caption{Impact of quantized point-wise features (\%).}
\label{tab:6}
\renewcommand\arraystretch{1.3}
\begin{tabular}{|c|c|c|} \hline
   Quantization & mIoU & OA \\ \hline \hline
   32-bit & 64.5 & 86.5 \\ \hline
   16-bit & 64.9 & 86.5 \\ \hline
\end{tabular}
\end{table}

\section{Conclusion and Future Work}
\label{sec:conclusion}

A green semantic segmentation method for large-scale indoor point clouds, called GSIP, was proposed in this work. It contains two novel ingredients: a new room-style method for data pre-processing and a new point cloud feature extractor which is extended from PointHop with lower memory and computational costs while preserving the segmentation performance. We evaluated the performance of GSIP against PointNet with the indoor S3DIS dataset and showed that GSIP outperforms PointNet in terms of performance accuracy, model sizes and computational complexity. As to future extension, it is desired to generalize the GSIP method from the large-scale indoor point clouds to the large-scale outdoor point clouds. The latter has many real world applications. Furthermore, the data imbalance problem should be carefully examined so as to boost the segmentation and/or classification performance. 

\bibliographystyle{unsrt}  
\bibliography{main}

\begin{thebibliography}{10}

\bibitem{qi2017pointnet}
Charles~R Qi, Hao Su, Kaichun Mo, and Leonidas~J Guibas.
\newblock Pointnet: Deep learning on point sets for 3d classification and
  segmentation.
\newblock In {\em Proceedings of the IEEE conference on computer vision and
  pattern recognition}, pages 652--660, 2017.

\bibitem{qi2017pointnet++}
Charles~R Qi, Li~Yi, Hao Su, and Leonidas~J Guibas.
\newblock Pointnet++: Deep hierarchical feature learning on point sets in a
  metric space.
\newblock {\em arXiv preprint arXiv:1706.02413}, 2017.

\bibitem{jiang2018pointsift}
Mingyang Jiang, Yiran Wu, Tianqi Zhao, Zelin Zhao, and Cewu Lu.
\newblock Pointsift: A sift-like network module for 3d point cloud semantic
  segmentation.
\newblock {\em arXiv preprint arXiv:1807.00652}, 2018.

\bibitem{li2018pointcnn}
Yangyan Li, Rui Bu, Mingchao Sun, Wei Wu, Xinhan Di, and Baoquan Chen.
\newblock Pointcnn: Convolution on x-transformed points.
\newblock {\em Advances in neural information processing systems}, 31:820--830,
  2018.

\bibitem{wang2019dynamic}
Yue Wang, Yongbin Sun, Ziwei Liu, Sanjay~E Sarma, Michael~M Bronstein, and
  Justin~M Solomon.
\newblock Dynamic graph cnn for learning on point clouds.
\newblock {\em Acm Transactions On Graphics (tog)}, 38(5):1--12, 2019.

\bibitem{wu20153d}
Zhirong Wu, Shuran Song, Aditya Khosla, Fisher Yu, Linguang Zhang, Xiaoou Tang,
  and Jianxiong Xiao.
\newblock 3d shapenets: A deep representation for volumetric shapes.
\newblock In {\em Proceedings of the IEEE conference on computer vision and
  pattern recognition}, pages 1912--1920, 2015.

\bibitem{armeni_cvpr16}
Iro Armeni, Ozan Sener, Amir~R. Zamir, Helen Jiang, Ioannis Brilakis, Martin
  Fischer, and Silvio Savarese.
\newblock 3d semantic parsing of large-scale indoor spaces.
\newblock In {\em Proceedings of the IEEE International Conference on Computer
  Vision and Pattern Recognition}, 2016.

\bibitem{hackel2017semantic3d}
Timo Hackel, Nikolay Savinov, Lubor Ladicky, Jan~D Wegner, Konrad Schindler,
  and Marc Pollefeys.
\newblock Semantic3d. net: A new large-scale point cloud classification
  benchmark.
\newblock {\em arXiv preprint arXiv:1704.03847}, 2017.

\bibitem{behley2019semantickitti}
Jens Behley, Martin Garbade, Andres Milioto, Jan Quenzel, Sven Behnke, Cyrill
  Stachniss, and Jurgen Gall.
\newblock Semantickitti: A dataset for semantic scene understanding of lidar
  sequences.
\newblock In {\em Proceedings of the IEEE/CVF International Conference on
  Computer Vision}, pages 9297--9307, 2019.

\bibitem{landrieu2018large}
Loic Landrieu and Martin Simonovsky.
\newblock Large-scale point cloud semantic segmentation with superpoint graphs.
\newblock In {\em Proceedings of the IEEE conference on computer vision and
  pattern recognition}, pages 4558--4567, 2018.

\bibitem{rethage2018fully}
Dario Rethage, Johanna Wald, Jurgen Sturm, Nassir Navab, and Federico Tombari.
\newblock Fully-convolutional point networks for large-scale point clouds.
\newblock In {\em Proceedings of the European Conference on Computer Vision
  (ECCV)}, pages 596--611, 2018.

\bibitem{guo2020pct}
Meng-Hao Guo, Jun-Xiong Cai, Zheng-Ning Liu, Tai-Jiang Mu, Ralph~R Martin, and
  Shi-Min Hu.
\newblock Pct: Point cloud transformer.
\newblock {\em arXiv preprint arXiv:2012.09688}, 2020.

\bibitem{hu2020randla}
Qingyong Hu, Bo~Yang, Linhai Xie, Stefano Rosa, Yulan Guo, Zhihua Wang, Niki
  Trigoni, and Andrew Markham.
\newblock Randla-net: Efficient semantic segmentation of large-scale point
  clouds.
\newblock In {\em Proceedings of the IEEE/CVF Conference on Computer Vision and
  Pattern Recognition}, pages 11108--11117, 2020.

\bibitem{zhang2020pointhop}
Min Zhang, Haoxuan You, Pranav Kadam, Shan Liu, and C-C~Jay Kuo.
\newblock Pointhop: An explainable machine learning method for point cloud
  classification.
\newblock {\em IEEE Transactions on Multimedia}, 22(7):1744--1755, 2020.

\bibitem{zhang2020pointhop++}
Min Zhang, Yifan Wang, Pranav Kadam, Shan Liu, and C-C~Jay Kuo.
\newblock Pointhop++: A lightweight learning model on point sets for 3d
  classification.
\newblock In {\em 2020 IEEE International Conference on Image Processing
  (ICIP)}, pages 3319--3323. IEEE, 2020.

\bibitem{zhang2020unsupervised}
Min Zhang, Pranav Kadam, Shan Liu, and C-C~Jay Kuo.
\newblock Unsupervised feedforward feature (uff) learning for point cloud
  classification and segmentation.
\newblock In {\em 2020 IEEE International Conference on Visual Communications
  and Image Processing (VCIP)}, pages 144--147. IEEE, 2020.

\bibitem{kadam2020unsupervised}
Pranav Kadam, Min Zhang, Shan Liu, and C-C~Jay Kuo.
\newblock Unsupervised point cloud registration via salient points analysis
  (spa).
\newblock In {\em 2020 IEEE International Conference on Visual Communications
  and Image Processing (VCIP)}, pages 5--8. IEEE, 2020.

\bibitem{kadam2021r}
Pranav Kadam, Min Zhang, Shan Liu, and C-C~Jay Kuo.
\newblock R-pointhop: A green, accurate and unsupervised point cloud
  registration method.
\newblock {\em arXiv preprint arXiv:2103.08129}, 2021.

\bibitem{chen2015xgboost}
Tianqi Chen, Tong He, Michael Benesty, Vadim Khotilovich, Yuan Tang, Hyunsu
  Cho, et~al.
\newblock Xgboost: extreme gradient boosting.
\newblock {\em R package version 0.4-2}, 1(4):1--4, 2015.

\bibitem{rusu2009fast}
Radu~Bogdan Rusu, Nico Blodow, and Michael Beetz.
\newblock Fast point feature histograms (fpfh) for 3d registration.
\newblock In {\em 2009 IEEE international conference on robotics and
  automation}, pages 3212--3217. IEEE, 2009.

\bibitem{tombari2010unique}
Federico Tombari, Samuele Salti, and Luigi Di~Stefano.
\newblock Unique signatures of histograms for local surface description.
\newblock In {\em European conference on computer vision}, pages 356--369.
  Springer, 2010.

\bibitem{hackel2016fast}
Timo Hackel, Jan~D Wegner, and Konrad Schindler.
\newblock Fast semantic segmentation of 3d point clouds with strongly varying
  density.
\newblock {\em ISPRS annals of the photogrammetry, remote sensing and spatial
  information sciences}, 3:177--184, 2016.

\bibitem{landrieu2017structured}
Loic Landrieu, Hugo Raguet, Bruno Vallet, Cl{\'e}ment Mallet, and Martin
  Weinmann.
\newblock A structured regularization framework for spatially smoothing
  semantic labelings of 3d point clouds.
\newblock {\em ISPRS Journal of Photogrammetry and Remote Sensing},
  132:102--118, 2017.

\bibitem{mallet2011relevance}
Cl{\'e}ment Mallet, Fr{\'e}d{\'e}ric Bretar, Michel Roux, Uwe Soergel, and
  Christian Heipke.
\newblock Relevance assessment of full-waveform lidar data for urban area
  classification.
\newblock {\em ISPRS journal of photogrammetry and remote sensing},
  66(6):S71--S84, 2011.

\bibitem{chehata2009airborne}
Nesrine Chehata, Li~Guo, and Cl{\'e}ment Mallet.
\newblock Airborne lidar feature selection for urban classification using
  random forests.
\newblock In {\em Laserscanning}, 2009.

\bibitem{lowe2004distinctive}
David~G Lowe.
\newblock Distinctive image features from scale-invariant keypoints.
\newblock {\em International journal of computer vision}, 60(2):91--110, 2004.

\bibitem{kuo2016understanding}
C-C~Jay Kuo.
\newblock Understanding convolutional neural networks with a mathematical
  model.
\newblock {\em Journal of Visual Communication and Image Representation},
  41:406--413, 2016.

\bibitem{kuo2018data}
C-C~Jay Kuo and Yueru Chen.
\newblock On data-driven saak transform.
\newblock {\em Journal of Visual Communication and Image Representation},
  50:237--246, 2018.

\bibitem{kuo2019interpretable}
C-C~Jay Kuo, Min Zhang, Siyang Li, Jiali Duan, and Yueru Chen.
\newblock Interpretable convolutional neural networks via feedforward design.
\newblock {\em Journal of Visual Communication and Image Representation},
  60:346--359, 2019.

\bibitem{chen2018saak}
Yueru Chen, Zhuwei Xu, Shanshan Cai, Yujian Lang, and C-C~Jay Kuo.
\newblock A saak transform approach to efficient, scalable and robust
  handwritten digits recognition.
\newblock In {\em 2018 Picture Coding Symposium (PCS)}, pages 174--178. IEEE,
  2018.

\bibitem{chen2020pixelhop}
Yueru Chen and C-C~Jay Kuo.
\newblock Pixelhop: A successive subspace learning (ssl) method for object
  recognition.
\newblock {\em Journal of Visual Communication and Image Representation},
  70:102749, 2020.

\bibitem{chen2020pixelhop++}
Yueru Chen, Mozhdeh Rouhsedaghat, Suya You, Raghuveer Rao, and C-C~Jay Kuo.
\newblock Pixelhop++: A small successive-subspace-learning-based (ssl-based)
  model for image classification.
\newblock In {\em 2020 IEEE International Conference on Image Processing
  (ICIP)}, pages 3294--3298. IEEE, 2020.

\bibitem{yang2021pixelhop}
Yijing Yang, Vasileios Magoulianitis, and C-C~Jay Kuo.
\newblock E-pixelhop: An enhanced pixelhop method for object classification.
\newblock {\em arXiv preprint arXiv:2107.02966}, 2021.

\bibitem{rouhsedaghat2020facehop}
Mozhdeh Rouhsedaghat, Yifan Wang, Xiou Ge, Shuowen Hu, Suya You, and C-C~Jay
  Kuo.
\newblock Facehop: A light-weight low-resolution face gender classification
  method.
\newblock {\em arXiv preprint arXiv:2007.09510}, 2020.

\bibitem{chen2021defakehop}
Hong-Shuo Chen, Mozhdeh Rouhsedaghat, Hamza Ghani, Shuowen Hu, Suya You, and
  C-C~Jay Kuo.
\newblock Defakehop: A light-weight high-performance deepfake detector.
\newblock In {\em 2021 IEEE International Conference on Multimedia and Expo
  (ICME)}, pages 1--6. IEEE, 2021.

\bibitem{zhang2021anomalyhop}
Kaitai Zhang, Bin Wang, Wei Wang, Fahad Sohrab, Moncef Gabbouj, and C-C~Jay
  Kuo.
\newblock Anomalyhop: An ssl-based image anomaly localization method.
\newblock {\em arXiv preprint arXiv:2105.03797}, 2021.

\bibitem{armeni2017joint}
Iro Armeni, Sasha Sax, Amir~R Zamir, and Silvio Savarese.
\newblock Joint 2d-3d-semantic data for indoor scene understanding.
\newblock {\em arXiv preprint arXiv:1702.01105}, 2017.

\bibitem{wold1987principal}
Svante Wold, Kim Esbensen, and Paul Geladi.
\newblock Principal component analysis.
\newblock {\em Chemometrics and intelligent laboratory systems}, 2(1-3):37--52,
  1987.

\end{thebibliography}

\end{document}